\documentclass[10pt,twocolumn,letterpaper]{article}
\usepackage[pagenumbers]{cvpr}
\usepackage{cuted}
\usepackage[nopatch=eqnum]{microtype}
\usepackage{bm}

\usepackage{amsmath,amsfonts,bm}

\def\eqref#1{equation~\ref{#1}}

\def\1{\bm{1}}

\DeclareMathAlphabet{\mathsfit}{\encodingdefault}{\sfdefault}{m}{sl}
\SetMathAlphabet{\mathsfit}{bold}{\encodingdefault}{\sfdefault}{bx}{n}

\usepackage{amssymb}
\usepackage{booktabs}
\usepackage{array}
\usepackage{colortbl}
\usepackage{graphicx}
\usepackage{url}
\usepackage{float}
\usepackage{wrapfig}
\usepackage{placeins}
\usepackage{color}
\usepackage{CJKutf8}

\definecolor{cvprblue}{rgb}{0.21,0.49,0.74}
\usepackage[pagebackref,breaklinks,colorlinks,allcolors=cvprblue]{hyperref}
\def\confName{CVPR}
\def\confYear{2027}

\title{Recovering the View: Benchmarking Physical Active Vision for Occlusion Recovery in Robotic Manipulation}
\author{
\normalsize
\textbf{Kaijun Luo}\textsuperscript{1} \quad
\textbf{Yudi Huang}\textsuperscript{2} \quad
\textbf{Qijun Zhong}\textsuperscript{1} \quad
\textbf{Xinshuai Song}\textsuperscript{1} \quad
\textbf{Yang Liu}\textsuperscript{1,4\ensuremath{\dagger}} \quad
\textbf{Liang Lin}\textsuperscript{1,3,4} \\
\small $^{1}$Sun Yat-sen University \quad
$^{2}$University of Electronic Science and Technology of China \\
\small $^{3}$Pengcheng Laboratory \quad $^{4}$X-Era AI Lab \\
\tt\footnotesize \{luokj6,zhongqj7,songxsh\}@mail2.sysu.edu.cn \quad huangyd@std.uestc.edu.cn \\
\tt\footnotesize liuy856@mail.sysu.edu.cn \quad linliang@ieee.org \\
\small $^{\dagger}$Corresponding author
}

\begin{document}
\maketitle

\setlength{\stripsep}{15pt}
\preCutedStrip{\vskip-25pt}
\postCutedStrip{\vskip10pt}
\begin{strip}
    \centering
    \includegraphics[width=1.0\textwidth]{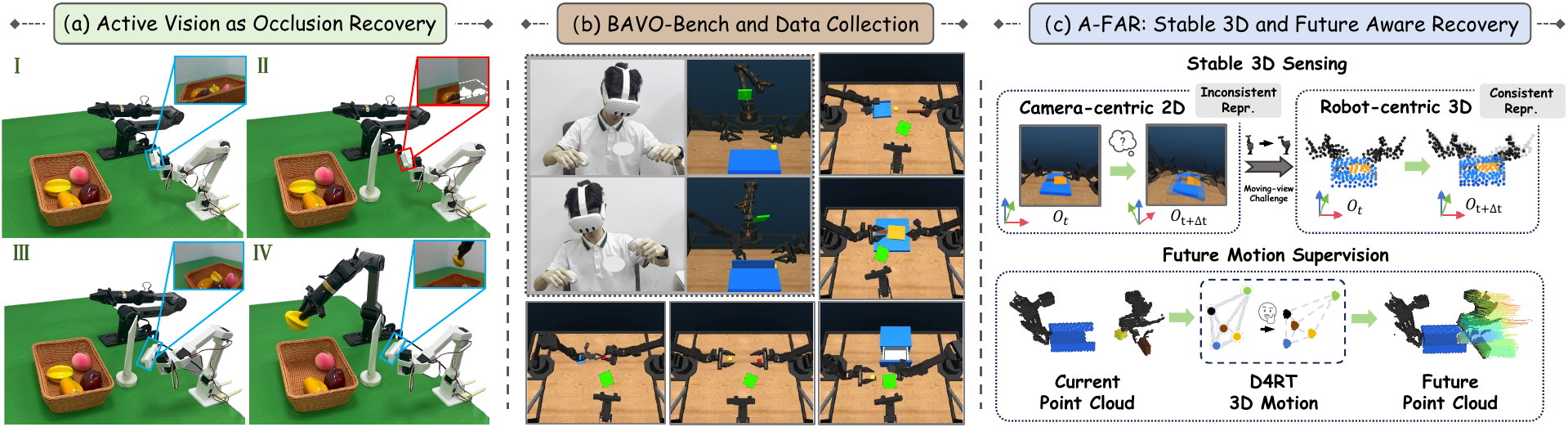}
    \captionsetup{hypcap=false}
    \captionof{figure}{\textbf{Overview of our motivation, benchmark, and approach.} \textbf{(a)} We study physical active vision as closed-loop recovery from external occlusion: using a single active camera as the sole visual sensor, the robot actively changes its viewpoint to recover task-relevant visibility and continue manipulation. \textbf{(b)} We introduce \textbf{BAVO-Bench}, a bimanual active-vision manipulation benchmark that systematically introduces controlled external occlusions to study viewpoint recovery, together with a VR teleoperation pipeline for collecting active-view demonstrations. \textbf{(c)} We present \textbf{A-FAR}, an active-vision policy that combines viewpoint-stable robot-centric 3D observations with future 3D motion supervision, enabling adaptive viewpoint and manipulation control under occlusion.}
    \label{fig:paper_overview}
\end{strip}
\preCutedStrip{}
\postCutedStrip{}

\begin{abstract}
Physical active vision allows robots to change their viewpoint when task-relevant observations become unreliable. However, existing manipulation benchmarks provide limited support for studying how policies recover from occlusion during execution. We introduce \textbf{BAVO-Bench}, a bimanual active-vision benchmark that systematically controls external visibility through Clean, Stage Occlusion, and Random-time Occlusion conditions. We further introduce \textbf{Camera-attributable Visibility Gain (CVG)}, a counterfactual metric that isolates visibility improvements attributable to camera motion rather than scene dynamics. Building on this setting, we present \textbf{A-FAR}, a policy for joint viewpoint and manipulation control. A-FAR combines robot-centric 3D observations with spatial context encoding to model geometric relationships consistently across viewpoints. It further leverages future 3D motion supervision from a pretrained 4D model to capture how task-relevant scene geometry evolves during manipulation. Experiments in simulation and the real world demonstrate improved robustness to structured and temporally unexpected occlusions, while maintaining strong performance under clean observations. Further analyses reveal that recovering visibility alone does not guarantee task success, highlighting the importance of translating recovered visual information into effective manipulation. Code and visualizations are available at: \href{https://hcplab-sysu.github.io/BAVO-Bench/}{BAVO-Bench}.
\end{abstract}

\section{Introduction}
\label{sec:introduction}

Robotic manipulation depends not only on how to act, but also on what the robot can see. Modern robot policies now range from task-specific visuomotor policies to generalist Vision-Language-Action (VLA) models and emerging World Action Models (WAMs), yet most still rely on passive visual observations from fixed or wrist-mounted cameras~\citep{zhao2023act,chi2023diffusion,kim2024openvla,black2024pi0,black2025pi05,ye2026dreamzero}. Such configurations can become unreliable when task-relevant regions are occluded or fall outside the available viewpoints, while wrist-mounted cameras further couple perception with manipulation motion. Recent methods have explored computational alternatives that select or synthesize virtual observations from existing scene evidence~\citep{liu2026activevla,heo2026anycamvla}. However, their effectiveness depends on the fidelity of the reconstructed or generated scene, particularly when task-relevant regions are heavily occluded or poorly observed. Physical active vision instead moves the sensor to a new line of sight, allowing the robot to acquire fresh measurements and actively restore task-relevant visibility~\citep{chuang2025active,xiong2025vision}.

Despite this progress, existing active-vision manipulation largely studies viewpoint requirements induced by task geometry or execution stages. For example, the robot may need to change its viewpoint to reveal an occluded target or acquire task-relevant visual evidence as execution progresses~\citep{wang2024observe,li2026activearena}. Prior work has further extended active-perception evaluation to occlusion-aware manipulation settings~\citep{he2026efm,liu2026sapave,cheng2018active}. Yet task-relevant visibility can also be disrupted unexpectedly during execution, for example when a human hand or a moved object blocks the current view. The policy must then recover an informative viewpoint before continuing manipulation. We therefore study physical active vision as closed-loop perceptual recovery from externally induced visibility failures during execution (Fig.~\ref{fig:paper_overview}(a)).

To systematically evaluate this capability, we introduce \textbf{BAVO-Bench} (\textbf{B}imanual \textbf{A}ctive \textbf{V}ision under \textbf{O}cclusion), a benchmark for bimanual manipulation with physical active vision built on the AV-ALOHA simulation platform~\citep{chuang2025active}. BAVO-Bench contains five multi-stage bimanual tasks and treats external occlusion as an explicit, controllable intervention while preserving the underlying task and control interface. Each task is evaluated under three visibility conditions: Clean, without artificial occlusion; Stage Occlusion, where task-relevant occlusions are introduced at structured stages and used for training demonstrations; and Random-time Occlusion, where similar visibility failures occur at temporally unexpected moments during execution. This design separates nominal manipulation, recovery under the demonstration distribution, and generalization to unexpected occlusion timing. Since task success alone cannot determine whether the policy actually recovered a useful view, we evaluate task progress and introduce \textbf{Camera-attributable Visibility Gain (CVG)} to measure whether camera motion itself restores task-relevant visibility (Fig.~\ref{fig:paper_overview}(b)).

Physical active vision also changes the learning problem itself. Camera motion can substantially change the observation even when the underlying scene remains unchanged. We therefore represent RGB-D observations as colored point clouds in a shared robot-base frame and build our policy on ManiFlow~\citep{yan2025maniflow} to obtain a more stable representation under viewpoint changes. A further challenge is that the policy also controls the sensor that generates its future observations. While viewpoint actions change how the scene is observed, manipulation actions can alter the underlying scene geometry. To capture how spatial structure remains stable or evolves under these coupled actions, we introduce \textbf{A-FAR} (\textbf{A}ctive \textbf{F}uture-\textbf{A}ware \textbf{R}ecovery), which combines spatial context modeling with future 3D motion supervision from D4RT~\citep{zhang2026d4rt}. By predicting action-conditioned point displacements, A-FAR encourages visual representations to encode scene dynamics while preserving geometric consistency across viewpoint changes. We further employ proprioceptive feature masking to promote observation-driven viewpoint control. The auxiliary motion supervision is used only during training, without requiring future information at inference (Fig.~\ref{fig:paper_overview}(c)).

In summary, our contributions are threefold:

\begin{itemize}
    \item We introduce \textbf{BAVO-Bench}, a five-task bimanual active-vision benchmark that studies recovery from externally induced occlusions under controlled intervention timing. We further propose \textbf{CVG}, a counterfactual metric that isolates visibility improvements attributable to camera motion.
    \item We propose \textbf{A-FAR}, an active-vision policy that combines robot-centric 3D observations and spatial context encoding with future 3D motion supervision, enabling viewpoint-stable representations and capturing how scene geometry evolves during manipulation.
    \item Extensive simulation and real-world experiments demonstrate that A-FAR improves manipulation performance and active-view recovery over representative baselines across diverse visibility conditions.
\end{itemize}
\section{Related Work}
\label{sec:related_work}

\noindent\textbf{Physical active vision for robotic manipulation.}
Active perception enables robots to change what they observe during manipulation. Early work jointly learned camera and manipulation actions under occlusion~\citep{cheng2018active}, while recent systems learn coordinated viewpoint control through movable sensors, including AV-ALOHA~\citep{chuang2025active}, Vision in Action~\citep{xiong2025vision}, and SaPaVe~\citep{liu2026sapave}. Most recently, ActiveScale scales this capability across model, data, and hardware through pose-grounded temporal modeling, large-scale egocentric training, and an independently actuated camera platform~\citep{zhou2026activescale}. Computational approaches such as ActiveVLA instead select informative viewpoints from reconstructed scene representations~\citep{liu2026activevla}. Our work focuses on physical active vision as \emph{closed-loop recovery}: restoring task-relevant visibility after it is externally disrupted during ongoing manipulation.

\noindent\textbf{Benchmarks for active-view manipulation.}
Recent benchmarks increasingly make information acquisition an explicit manipulation capability. EFM-10 organizes tasks around exploratory and focused perception~\citep{he2026efm}, ActiveManip-Bench evaluates unoccluded, occluded, and out-of-view conditions~\citep{liu2026sapave}, and TAVIS studies active sensing under procedural shifts and evaluates anticipatory gaze~\citep{spigler2026tavis}. In contrast, we treat visibility failure as a controllable \emph{within-episode intervention} over an otherwise unchanged task. Stage Occlusion evaluates demonstrated recovery structure, while Random-time Occlusion breaks the correlation between task progress and intervention onset. Our CVG metric further compares the recovered view with a same-state frozen-camera counterfactual, attributing visibility improvement specifically to camera motion.

\setcounter{figure}{1}
\begin{figure*}[t]
    \centering
    \includegraphics[width=0.9\textwidth]{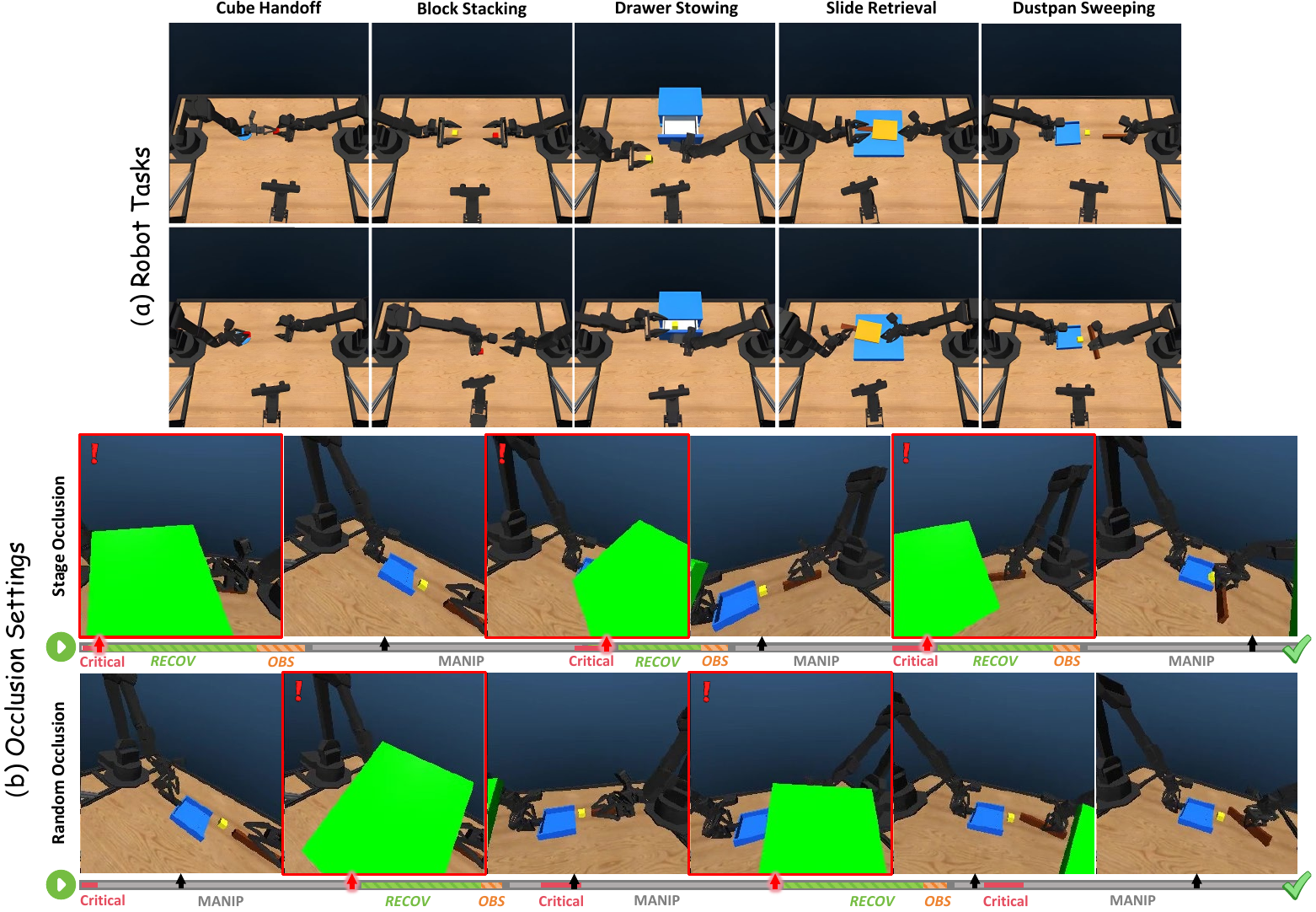}
    \vspace{-10pt}
    \caption{\textbf{BAVO-Bench overview.} \textbf{Top:} Five multi-stage bimanual manipulation tasks. \textbf{Bottom:} Representative rollouts under the two occlusion settings. Red-bordered frames mark occlusion events, and the timelines illustrate recovery and manipulation behaviors. Both settings occlude task-relevant targets; \textbf{Stage Occlusion} aligns interventions with predefined task transitions, whereas \textbf{Random-time Occlusion} samples intervention times independently of these transitions.}
     \vspace{-10pt}
    \label{fig:benchmark_overview}
\end{figure*}

\noindent\textbf{Viewpoint-stable and future-aware representations.}
Active camera motion introduces viewpoint-dependent changes in visual observations while coupling camera actions with future sensory inputs. ActiveScale addresses viewpoint robustness using temporal images and explicit camera-pose supervision~\citep{zhou2026activescale}, while EgoAVFlow jointly predicts future 3D flow and camera trajectories for active vision~\citep{cho2026egoavflow}. MECo-WAM distills temporal 4D geometric relations into an inference-efficient world-action policy~\citep{zhang2026mecowam}. In contrast, A-FAR combines viewpoint-stable 3D observations with D4RT-derived pointwise motion supervision~\citep{zhang2026d4rt}, encouraging visual representations to capture geometric stability and dynamics under joint viewpoint and manipulation actions. This auxiliary supervision is applied only during training, without requiring future information or motion prediction at inference.

\section{BAVO-Bench: Physical Active-View Recovery under External Occlusion}
\label{sec:benchmark}

We build \textbf{BAVO-Bench} on the AV-ALOHA active-vision simulator~\citep{chuang2025active}. The original AV-ALOHA benchmark primarily studies tasks whose viewpoint requirements arise from task geometry or predictable execution stages. To focus on recovery from externally induced visibility loss, we design five multi-stage tasks around common bimanual manipulation patterns and introduce external occlusion as a controlled intervention, as illustrated in Fig.~\ref{fig:benchmark_overview}.

\subsection{Controlled Visibility Interventions}
\label{sec:benchmark_interventions}

The three conditions differ in whether and when external occlusion is introduced. At each intervention, a task-relevant target is selected according to task progress, and a camera-facing occluder is placed along its line of sight from the active camera. Placements are validated for target occlusion and robot clearance, and the occluder remains fixed in the world frame until the next intervention. Task definitions and additional intervention details are provided in Appendix~\ref{app:benchmark_details}.

\noindent\textbf{Clean.}
Clean denotes nominal task execution without benchmark occlusion. It serves as the reference for the policy's manipulation performance under standard visual conditions.

\noindent\textbf{Stage Occlusion.}
Each episode begins with an initial task-relevant occlusion. As execution progresses, the occlusion target is updated at two predefined task transitions, producing three structured occlusion phases over the trajectory. Because these transitions are fixed across demonstrations, Stage provides the training-matched setting for learning and evaluating active-view recovery.

\noindent\textbf{Random-time Occlusion.}
Each episode likewise begins with an initial occlusion, but subsequent interventions are sampled from predefined temporal windows independently of task transitions. Unlike Stage Occlusion, which introduces two stage-aligned updates, Random-time schedules up to six additional interventions that may interrupt ongoing manipulation. This setting jointly challenges recovery from unexpected occlusion timing and more frequent visibility disruptions, rather than isolating intervention timing alone.

\subsection{Demonstration and Evaluation Protocol}
\label{sec:benchmark_protocol}

We collect 100 VR demonstrations per task under Stage Occlusion, yielding 500 trajectories in total. Following a new occlusion, operators are encouraged to primarily reposition the observation arm while keeping manipulation approximately stable; once an informative viewpoint is recovered, the observation arm remains comparatively stable and task execution continues. This provides structured demonstrations of physical view recovery followed by manipulation.

Policies are trained exclusively on Stage demonstrations and evaluated using the same trained model under all three conditions. This protocol evaluates nominal manipulation under Clean, training-matched recovery under Stage, and recovery under unscheduled and potentially more frequent visibility disruptions.

\subsection{Evaluation Metrics}
\label{sec:benchmark_metrics}

We report Success Rate (SR) and Task Progress Score (TPS) for task performance, together with Camera-attributable Visibility Gain (CVG) to directly evaluate visibility recovery attributable to viewpoint change.

\noindent\textbf{Success Rate and Task Progress.}
Success Rate (SR) measures the fraction of episodes that complete the full task. To capture partial completion, we additionally report Task Progress Score (TPS). Each task contains five ordered milestones; if $d_i\in\{0,\ldots,5\}$ denotes the deepest milestone reached in episode $i$, then $\mathrm{TPS}=\frac{1}{N}\sum_{i=1}^{N} d_i/5$. Together, SR and TPS characterize task performance at complementary granularities: final task completion and intermediate progress.

\noindent\textbf{Camera-attributable Visibility Gain.}
To isolate the sensing consequence of viewpoint change, we introduce Camera-attributable Visibility Gain (CVG). Consider an occlusion event $e$ introduced at time $\tau_e$, and let $x_t$ denote the current robot and object configuration. For an evaluation camera pose $C$, we measure normalized target visibility as
\begin{equation}
V_e(C,x_t) = \frac{N^{\mathrm{occ}}_e(C,x_t)}
{N^{\mathrm{noOcc}}_e(C,x_t)},
\label{eq:visibility}
\end{equation}
where $N^{\mathrm{occ}}_e$ and $N^{\mathrm{noOcc}}_e$ are the visible target pixels rendered from the same camera and scene state with and without the benchmark occluder, respectively. The ratio therefore measures how much of the target that would otherwise be visible remains visible under the external occlusion.

CVG compares the policy's realized camera trajectory with a counterfactual reference in which the camera remains fixed at its event-start pose $C_{\tau_e}$. Over a recovery window of $H$ steps,
\begin{equation}
\mathrm{CVG}_e = \frac{1}{H}
\sum_{h=0}^{H-1}
\left[
V_e(C_{\tau_e+h},x_{\tau_e+h})
- V_e(C_{\tau_e},x_{\tau_e+h})
\right].
\label{eq:event_cvg}
\end{equation}

Crucially, both visibility terms are evaluated on the same current scene state $x_{\tau_e+h}$. Changes caused by manipulation or object motion are therefore shared by both views, while their difference captures the visibility advantage associated with viewpoint change. A positive CVG indicates that the policy-selected camera trajectory reveals more of the task-relevant target than would have been visible had the camera remained at its event-start pose.

The counterfactual renders are used only for privileged simulation-time evaluation and never affect policy observations or actions. Recovery-window and aggregation details for CVG evaluation are provided in Appendix~\ref{app:benchmark_details}.

\section{A-FAR: Active Future-Aware Recovery}
\label{sec:method}

We present \textbf{A-FAR}, a ManiFlow-based policy~\citep{yan2025maniflow} for joint manipulation and active-view control. As illustrated in Fig.~\ref{fig:afar_overview}, A-FAR combines viewpoint-stable 3D observations with future 3D motion supervision to learn future-aware visuomotor representations. By encouraging the policy to capture scene dynamics and rely on visual evidence, A-FAR enables more adaptive viewpoint control under occlusion.

\setcounter{figure}{2}
\begin{figure*}[t]
    \centering
    \includegraphics[width=0.8\textwidth]{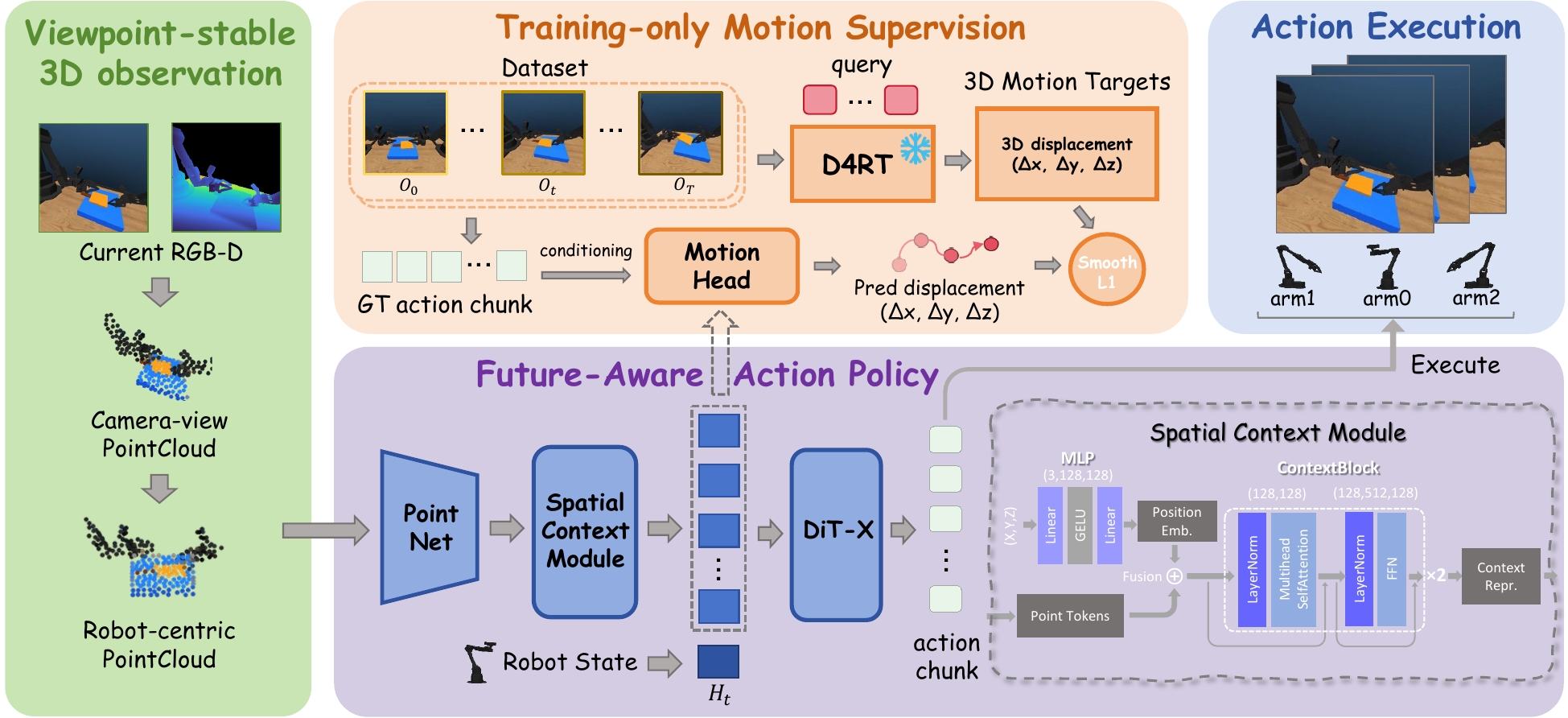}
    \caption{\textbf{Overview of A-FAR.} RGB-D observations are transformed into robot-base point clouds and encoded by PointNet and a spatial context module. The resulting point tokens, combined with robot states, condition ManiFlow's DiT-X to generate manipulation and viewpoint actions. During training, D4RT provides future 3D point displacement targets, while a motion head predicts these displacements from contextualized point features and demonstrated future actions. The D4RT teacher and auxiliary motion head are removed at inference.}
     \vspace{-10pt}
    \label{fig:afar_overview}
\end{figure*}

\subsection{Viewpoint-Stable Spatial Representation}
\label{sec:afar_observation}

Active-view control requires the policy to reason about scene changes despite continual camera motion. However, even static scene geometry can exhibit substantial variation in image space or camera-frame coordinates as the viewpoint changes. We therefore transform RGB-D observations into a shared robot-base frame to establish a consistent spatial reference.

Specifically, let $T_{B\leftarrow E_t}\in SE(3)$ denote the transformation from the observation-arm end-effector frame $E_t$ to the robot-base frame $B$, and $T_{E\leftarrow C}\in SE(3)$ the fixed camera-to-end-effector transformation. Their composition gives the camera-to-base transformation $T_{B\leftarrow C_t}=T_{B\leftarrow E_t}T_{E\leftarrow C}$. We back-project RGB-D observations into 3D and transform the resulting points into the robot-base frame, obtaining a colored point cloud $P_t\in\mathbb{R}^{N\times6}$. This shared frame provides consistent coordinates for static geometry across viewpoints while preserving changes in scene motion and visibility.

Beyond coordinate consistency, active-view control also requires modeling spatial relationships among scene elements, particularly between targets and occluders. Following ManiFlow~\citep{yan2025maniflow}, we first extract pointwise features using a PointNet without global pooling. Since these features are independently encoded without explicit inter-point interactions, we introduce a lightweight spatial context module that incorporates 3D positional embeddings and applies self-attention across point tokens. The resulting contextualized representations $H_t$ are used for subsequent policy learning.

\subsection{Future 3D Motion Supervision}
\label{sec:afar_motion}

Demonstration trajectories provide rich information about how scene geometry evolves during robot interactions. We exploit this temporal information as privileged supervision, encouraging the visual representations to capture action-dependent 3D motion that is relevant to manipulation and active-view control.

\noindent\textbf{Future motion targets.}
We adopt OpenD4RT~\citep{opend4rt}, an open-source implementation of D4RT~\citep{zhang2026d4rt}, to extract pointwise 3D motion from demonstration trajectories. For each sampled point in $P_t$, we retain its corresponding pixel coordinates $(u_i,v_i)$ in the original RGB-D observation before back-projection. Following D4RT's spatiotemporal query formulation, we construct $q_i=(u_i,v_i,t,t+\Delta t,t)$, specifying the source point, future target time, and reference camera frame. Given the demonstration clip and these queries, D4RT directly predicts a 3D displacement $\mathbf{d}_{t,i}^{\mathrm{D4RT}}$ for each point. We transform the predicted displacement into the robot-base frame:
\begin{equation}
\Delta\mathbf{x}_{t,i}^{T}
= R_{B\leftarrow C_t}
\mathbf{d}_{t,i}^{\mathrm{D4RT}},
\label{eq:afar_motion_target}
\end{equation}
where $R_{B\leftarrow C_t}$ is the rotational component of the camera-to-base transformation. Valid predictions are retained as motion targets and precomputed for auxiliary supervision during training.

\begin{table*}[t]
    \centering
    \small
    \caption{\textbf{Main results on BAVO-Bench.}
SR and TPS (\%) are macro-averaged across five tasks.
ACT, ManiFlow, and A-FAR report mean $\pm$
standard deviation over three training seeds,
while $\pi_{0.5}$ is reported from a single
multi-task fine-tuning run. Higher is better.}
 \vspace{-10pt}
    \label{tab:main_results}

    \setlength{\tabcolsep}{1pt}
    \renewcommand{\arraystretch}{0.95}
    \begin{tabular*}{\textwidth}{@{\extracolsep{\fill}}lcccccc@{}}
        \toprule
        & \multicolumn{2}{c}{Clean}
        & \multicolumn{2}{c}{Stage Occlusion}
        & \multicolumn{2}{c}{Random-time Occlusion} \\
        \cmidrule(lr){2-3}
        \cmidrule(lr){4-5}
        \cmidrule(lr){6-7}

        Method
        & {\footnotesize SR} & {\footnotesize TPS} 
        & {\footnotesize SR}  & {\footnotesize TPS} 
        & {\footnotesize SR}  & {\footnotesize TPS}  \\
        \midrule

        ACT
        & $10.3{\pm}1.2$ & $46.2{\pm}6.8$
        & $24.7{\pm}2.3$ & $58.1{\pm}1.7$
        & $15.7{\pm}3.2$ & $46.6{\pm}0.4$ \\

        $\pi_{0.5}$
        & 13.0 & 55.2
        & 42.0 & 73.4
        & 4.0 & 40.8 \\

        ManiFlow
        & $45.7{\pm}1.5$ & $70.5{\pm}6.4$
        & $60.7{\pm}3.1$ & $81.5{\pm}3.2$
        & $15.0{\pm}3.5$ & $50.5{\pm}2.7$ \\

        \textbf{A-FAR}
        & $\mathbf{50.0{\pm}2.6}$ & $\mathbf{75.5{\pm}1.6}$
        & $\mathbf{69.7{\pm}2.5}$ & $\mathbf{88.2{\pm}1.6}$
        & $\mathbf{20.3{\pm}3.1}$ & $\mathbf{52.1{\pm}1.8}$ \\

        \bottomrule
    \end{tabular*}
     \vspace{-10pt}
\end{table*}

\noindent\textbf{Action-conditioned motion prediction.}
Given the contextualized point tokens $H_t$ and the future action sequence $\mathbf{A}_t=(\mathbf{a}_t,\ldots,\mathbf{a}_{t+\Delta t-1})$ from demonstrations, a lightweight motion head $g_\psi$ predicts the 3D displacement of each point, $\widehat{\Delta X}_t=g_\psi(H_t,\mathbf{A}_t)$. Conditioning on the demonstrated actions allows the auxiliary task to associate visual features with action-dependent geometric changes. The predicted displacements are supervised using a Smooth L1 loss over valid points:
\begin{equation}
\mathcal{L}_{\mathrm{motion}}
= \frac{1}{\sum_i m_i}
\sum_i m_i\,
\operatorname{SmoothL1}
\left(
\widehat{\Delta\mathbf{x}}_{t,i},
\Delta\mathbf{x}_{t,i}^{T}
\right),
\label{eq:afar_motion_loss}
\end{equation}
where $m_i$ indicates the validity of each motion target. The auxiliary objective is jointly optimized with ManiFlow's flow matching and consistency losses:
\begin{equation}
\mathcal{L}
=
\mathcal{L}_{\mathrm{flow}}
+
\mathcal{L}_{\mathrm{cons}}
+
\lambda_{\mathrm{motion}}\mathcal{L}_{\mathrm{motion}},
\label{eq:afar_training_objective}
\end{equation}
where $\lambda_{\mathrm{motion}}$ controls the contribution of motion supervision. At inference, the motion head and all future supervision are removed.

\subsection{Vision-Grounded Action Learning}
\label{sec:afar_policy}

Learning motion-aware visual representations alone does not guarantee that the policy will rely on visual evidence for viewpoint control. In imitation learning, robot states can serve as shortcuts, allowing the policy to associate proprioceptive configurations with demonstrated camera motions while overlooking changes in scene occlusion. To encourage observation-driven viewpoint adaptation, we employ stochastic proprioceptive feature masking during training.

Specifically, the robot state $s_t$ is projected into an embedding $e_t$ through an MLP and concatenated with each contextualized point token $h_{t,i}$ from the visual encoder. During training, we randomly mask the entire state embedding with probability $p$, while preserving the visual features:
\begin{equation}
c_{t,i} = [h_{t,i};\, m e_t],
\qquad m\sim\operatorname{Bernoulli}(1-p).
\label{eq:afar_state_masking}
\end{equation}
The same mask is shared across all points and observations within each training window. The resulting tokens condition ManiFlow's DiT-X to generate both manipulation and viewpoint actions. At inference, the full state embedding is always retained.

\section{Experiments}
\label{sec:experiments}

We evaluate A-FAR on BAVO-Bench and a physical active-camera manipulation platform to address five core research questions: 
(\textbf{1}) How effectively does A-FAR handle nominal manipulation, structured occlusion recovery, and temporally unexpected occlusions compared with state-of-the-art visuomotor policies? (Sec.~\ref{sec:main_results})
(\textbf{2}) Does actively recovering target visibility directly translate into downstream task completion? (Sec.~\ref{sec:active_vision_analysis})
(\textbf{3}) How do robot-centric 3D alignment, spatial context modeling, future motion supervision, and proprioceptive masking contribute to active-view control? (Sec.~\ref{sec:ablation_studies})
(\textbf{4}) Do the learned visual representations capture richer spatial and temporal geometric information? (Sec.~\ref{sec:representation_analysis})
(\textbf{5}) Can A-FAR achieve closed-loop active-view recovery and manipulation on physical hardware under external occlusions? (Sec.~\ref{sec:real_world})

\subsection{Experimental Setup}

\noindent\textbf{Baselines.}
We compare A-FAR against three representative baselines spanning distinct visuomotor paradigms: 
(1) \textbf{ACT}~\cite{zhao2023act}, a camera-centric 2D action-chunking transformer policy; 
(2) $\bm{\pi_{0.5}}$~\cite{black2025pi05}, a large-scale pretrained Vision-Language-Action (VLA) generalist model operating on 2D images; and 
(3) \textbf{ManiFlow}~\cite{yan2025maniflow}, a 3D point-cloud consistency flow policy that serves as our architectural backbone.

\noindent\textbf{Training and evaluation.}
Following the BAVO-Bench protocol (Sec.~\ref{sec:benchmark_protocol}), all policies are trained exclusively on Stage Occlusion demonstrations and evaluated under Clean, Stage Occlusion, and Random-time Occlusion. ACT, ManiFlow, and A-FAR are trained per task across three random seeds, whereas $\pi_{0.5}$ is fine-tuned once across all five tasks due to its computational footprint. Each model checkpoint is evaluated for 20 episodes per task and condition. We report macro-averaged Success Rate (SR) and Task Progress Score (TPS) across the five tasks, alongside Camera-attributable Visibility Gain (CVG) under Random-time Occlusion. Unless otherwise noted, ablation studies use a single training seed and follow the same rollout evaluation protocol.

\begin{table*}[t]
    \centering
    \scriptsize
    \caption{\textbf{Active-view recovery under Random-time Occlusion.}
    CVG measures visibility gains attributable to camera motion. $\mathrm{Joint}=\mathrm{SR}\times\mathrm{CVG}/100$
weights CVG by task success.
Avg. denotes the mean across five tasks.
Higher is better.}
     \vspace{-10pt}
    \label{tab:active_view}

    \setlength{\tabcolsep}{2pt}
    \renewcommand{\arraystretch}{1.0}
    \begin{tabular*}{0.90\textwidth}{@{\extracolsep{\fill}}l>{\columncolor{black!8}}cccccc@{}}
        \toprule
        Method & Avg. & Cube Handoff & Block Stacking & Drawer Stowing & Slide Retrieval & Dustpan Sweeping \\
        \midrule
        \multicolumn{7}{l}{\textit{CVG (\%)}} \\
        ACT & 19.14 & 16.49 & 20.36 & 19.71 & 17.64 & 21.49 \\
        $\pi_{0.5}$ & 34.42 & \textbf{32.81} & 32.94 & 35.65 & 28.72 & 41.99 \\
        ManiFlow & 28.57 & 26.25 & 28.66 & 30.47 & 22.50 & 34.97 \\
        \textbf{A-FAR} & \textbf{34.77} & 30.51 & \textbf{33.54} & \textbf{36.36} & \textbf{30.73} & \textbf{42.71} \\
        \midrule
        \multicolumn{7}{l}{\textit{Joint (\%)}} \\
        ACT & 2.97 & 0.00 & 0.00 & 8.87 & 5.29 & 0.71 \\
        $\pi_{0.5}$ & 1.41 & 0.00 & 0.00 & 0.00 & 2.87 & 4.20 \\
        ManiFlow & 4.42 & 2.18 & 0.00 & 11.67 & 2.99 & 5.25 \\
        \textbf{A-FAR} & \textbf{7.35} & \textbf{3.05} & 0.00 & \textbf{15.15} & \textbf{7.17} & \textbf{11.39} \\
        \bottomrule
    \end{tabular*}
     \vspace{-10pt}
\end{table*}

\subsection{Main Simulation Results}
\label{sec:main_results}

\noindent\textbf{Overall manipulation performance.}
As shown in Table~\ref{tab:main_results}, A-FAR consistently achieves the highest average SR and TPS across all three visibility conditions, outperforming both image-based policies and its ManiFlow backbone. Under Stage Occlusion, A-FAR raises SR from 60.7\% to 69.7\% and TPS from 81.5\% to 88.2\% over ManiFlow by tightly coordinating viewpoint adjustments with multi-stage manipulation. Notably, although all policies are trained exclusively on Stage demonstrations, A-FAR transfers most effectively to unoccluded Clean execution, improving SR from 45.7\% to 50.0\% over ManiFlow without overfitting to the demonstrated occlusion-recovery routine.

\noindent\textbf{Adaptation to unexpected visibility disruptions.}
Random-time Occlusion introduces unscheduled and potentially more frequent occlusions that may interrupt ongoing manipulation. All methods experience substantial performance degradation, particularly the pretrained $\pi_{0.5}$, whose SR drops from 42.0\% under Stage Occlusion to 4.0\%. In contrast, A-FAR achieves the highest average SR of 20.3\%, representing a 35\% relative improvement over ManiFlow. These results highlight A-FAR's improved ability to adapt joint viewpoint and manipulation control to visibility disruptions beyond the predefined recovery stages encountered during training.

\subsection{Active Vision Analysis}
\label{sec:active_vision_analysis}

\noindent\textbf{Camera-attributable visibility recovery.}
While SR and TPS characterize manipulation outcomes, CVG directly evaluates the visibility gained through active camera motion. As shown in Table~\ref{tab:active_view}, A-FAR achieves the highest average CVG of 34.77\%, compared with 28.57\% for ManiFlow, with improvements across all five tasks. Because CVG compares the realized camera trajectory against a frozen-camera counterfactual at the exact same scene state, this margin confirms that A-FAR actively repositions the sensor to clear line-of-sight obstructions rather than relying on incidental object motion.

\noindent\textbf{The visibility--execution gap.}
Interestingly, $\pi_{0.5}$ achieves a comparable CVG of 34.42\%, yet its Random-time SR remains at only 4.0\%. This discrepancy exposes a pronounced visibility--execution gap, where actively restoring target visibility does not automatically translate into successful manipulation. To evaluate whether recovered views actually contribute to task completion, we report $\mathrm{Joint}=\mathrm{SR}\times\mathrm{CVG}/100$, which weights visibility gain by task success. Under this metric, precision-critical Block Stacking yields 0.00\% across all methods: although policies recover substantial target visibility (e.g., 33.54\% CVG for A-FAR), repeated mid-trajectory interruptions still disrupt fine-grained alignment. Across all tasks, however, A-FAR achieves the highest average Joint score of 7.35\%, outperforming both ManiFlow (4.42\%) and $\pi_{0.5}$ (1.41\%) by more effectively translating recovered visual evidence into closed-loop motor control.

\subsection{Ablation Studies}
\label{sec:ablation_studies}

\begin{table}[t]
    \centering
    \footnotesize
    \caption{\textbf{Ablation study of A-FAR.}
SR (\%) is reported under Clean, Stage,
and Random-time Occlusion, with CVG evaluated
under Random-time Occlusion. All variants use a
single training seed. 
Base and camera indicate the coordinate frames
used for point-cloud representation.}
\vspace{-10pt}
    \label{tab:ablation}

    \setlength{\tabcolsep}{3pt}
    \renewcommand{\arraystretch}{0.9}
    \begin{tabular*}{\columnwidth}{@{\extracolsep{\fill}}lcccc@{}}
        \toprule
        Variant & Clean & Stage & Random & CVG \\
        \midrule
        \multicolumn{5}{l}{
            \textit{Component ablation (base frame)}
        } \\
        \midrule
        ManiFlow
        & 44 & 60 & 19 & 30.3 \\

        + Context
        & 43 & 61 & 17 & 29.7 \\

        + Context + Motion
        & 48 & \textbf{73} & 19 & 31.1 \\

        + Motion + Mask
        & 51 & 64 & 18 & 34.3 \\

        \textbf{A-FAR (full)}
        & \textbf{52} & 72
        & \textbf{23} & \textbf{34.6} \\

        \midrule
        \multicolumn{5}{l}{
            \textit{Coordinate frame ablation}
        } \\
        \midrule
        ManiFlow (camera)
        & 26 & 44 & 14 & 27.1 \\

        A-FAR (camera)
        & 29 & 56 & 11 & 33.1 \\
        \bottomrule
    \end{tabular*}
    \vspace{-10pt}
\end{table}

\noindent\textbf{Complementary contributions of A-FAR components.}
Table~\ref{tab:ablation} examines the contributions of spatial context encoding, future motion supervision, and proprioceptive masking. Adding spatial context alone yields limited gains, as sparse action supervision is insufficient to constrain point-to-point geometric interactions. Pairing spatial context with dense motion supervision, however, increases Stage SR from 61\% to 73\% and Clean SR from 43\% to 48\%. Conversely, removing spatial context from the full model reduces Stage SR from 72\% to 64\% and Random-time SR from 23\% to 18\%, confirming that relational point encoding becomes essential once guided by dense 3D motion targets. Finally, incorporating proprioceptive masking improves Random-time SR from 19\% to 23\% and raises CVG from 31.1\% to 34.6\%, showing that suppressing state shortcuts promotes observation-driven viewpoint recovery under unexpected occlusions.

\noindent\textbf{Importance of robot-centric coordinates.}
We further investigate the role of a shared robot-base frame by comparing it against representations expressed in the moving camera frame. Using camera-frame coordinates consistently degrades manipulation performance for both ManiFlow and A-FAR. For A-FAR, Clean SR drops from 52\% to 29\%, while Random-time SR is reduced by more than half. In contrast, CVG remains at 33.1\%, close to the 34.6\% obtained with base-frame coordinates, despite the substantial reduction in task success. This divergence indicates that even when the camera successfully reveals the target, the policy cannot reliably translate newly observed geometry into accurate manipulation actions without a stationary spatial reference.

\subsection{Representation Analysis}
\label{sec:representation_analysis}

\begin{table}[t]
    \centering
    \footnotesize
    \caption{\textbf{Future visibility probing.}
    Test AUROC for predicting point visibility
    after 15 steps using frozen visual features
    without future actions. Higher is better.}
    \vspace{-10pt}
    \label{tab:visibility_probe}
    \setlength{\tabcolsep}{3pt}
    \renewcommand{\arraystretch}{0.9}
    \begin{tabular}{@{}lcccccc@{}}
        \toprule
        Method & Cube & Stack & Drawer & Slide & Sweep & Avg. \\
        \midrule
        ManiFlow & 0.6751 & 0.6652 & 0.7012 & 0.6717 & 0.6940 & 0.6815 \\
        \textbf{A-FAR} & \textbf{0.8191} & \textbf{0.8157} & \textbf{0.8391} & \textbf{0.8387} & \textbf{0.8701} & \textbf{0.8365} \\
        \bottomrule
    \end{tabular}
    \vspace{-10pt}
\end{table}

\begin{figure*}[t]
    \centering
    \includegraphics[
        width=\textwidth,
        keepaspectratio
    ]{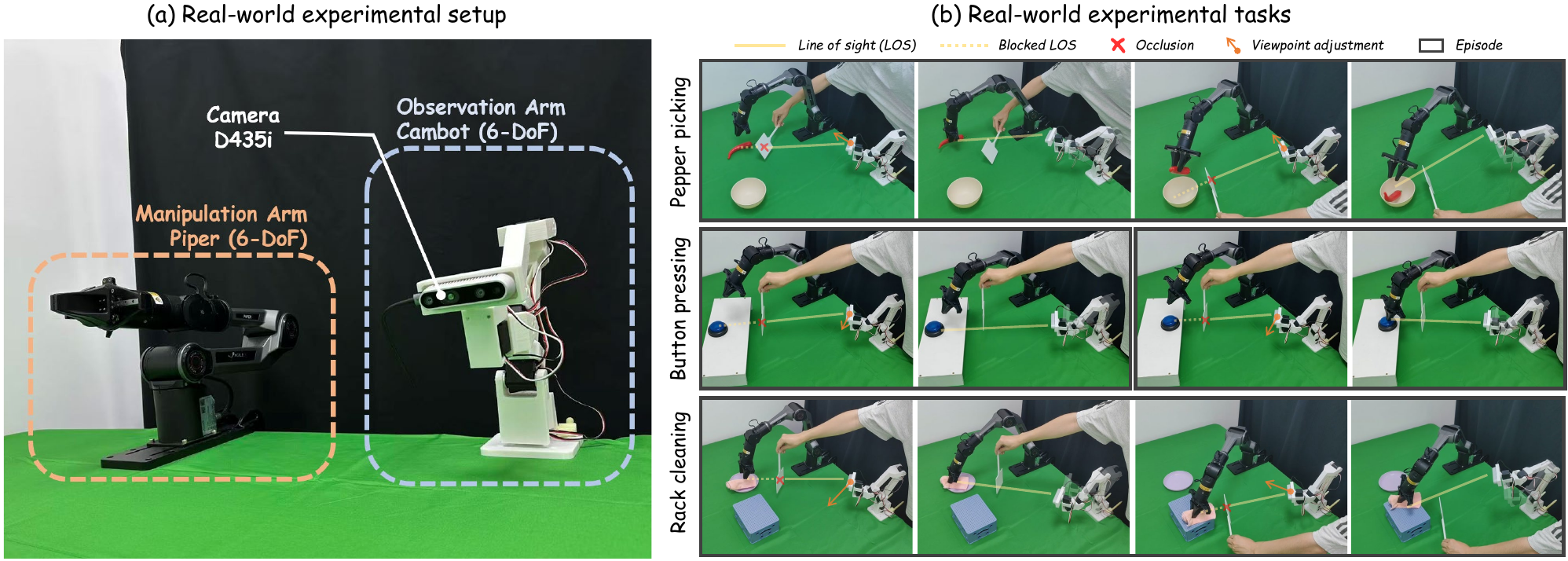}
    \caption{\textbf{Real-world experimental platform and tasks.}
(a) Our active-vision platform consists of a 6-DoF observation arm equipped with a D435i camera and a 6-DoF Piper manipulation arm.
(b) Representative execution sequences of Pepper Picking, Button Pressing, and Rack Cleaning under human-introduced occlusions, illustrating active viewpoint adjustment and manipulation.}
    \label{fig:real_robot_setup}
\end{figure*}

To test whether A-FAR encodes 3D spatial structure and scene dynamics, we probe future point visibility from frozen visual representations. Specifically, we freeze the visual encoders and train a lightweight MLP to predict whether each observed 3D point remains visible after 15 steps using point-wise features alone, without future actions. Ground-truth binary visibility labels at $t+15$ are obtained directly from the simulator. For each task, we split the 100 Stage Occlusion demonstrations into 80 training and 20 test episodes, sampling 16 frames per episode.

As shown in Table~\ref{tab:visibility_probe}, A-FAR consistently outperforms ManiFlow across all five tasks, improving the average AUROC from 0.6815 to 0.8365. Although A-FAR is supervised with 3D point displacements rather than explicit visibility labels during policy training, its frozen point tokens allow accurate decoding of upcoming occlusions, indicating that joint context attention and motion supervision embed predictive spatiotemporal geometry.

\subsection{Real-World Experiments}
\label{sec:real_world}

\begin{table}[t]
    \centering
    \footnotesize
    \caption{\textbf{Real-world manipulation results.}
    SR (\%) under Clean and Stage Occlusion, with 10 trials per method, task, and condition.}
    \vspace{-10pt}
    \label{tab:real_world_results}
    \setlength{\tabcolsep}{3pt}
    \renewcommand{\arraystretch}{0.9}
    \begin{tabular*}{0.92\columnwidth}{@{\extracolsep{\fill}}lcccccc@{}}
        \toprule
        & \multicolumn{2}{c}{Button}
        & \multicolumn{2}{c}{Pepper}
        & \multicolumn{2}{c}{Rack} \\
        \cmidrule(lr){2-3}
        \cmidrule(lr){4-5}
        \cmidrule(lr){6-7}
        Method
        & {\scriptsize Clean} & {\scriptsize Stage}
        & {\scriptsize Clean} & {\scriptsize Stage}
        & {\scriptsize Clean} & {\scriptsize Stage} \\
        \midrule
        ACT & 60 & 40 & 50 & 20 & 50 & 60 \\
        $\pi_{0.5}$ & 50 & 50 & \textbf{70} & 40 & 60 & 40 \\
        ManiFlow & 60 & 50 & 50 & 20 & 60 & 30 \\
        \textbf{A-FAR} & \textbf{70} & \textbf{90} & 60 & \textbf{60} & \textbf{70} & \textbf{70} \\
        \bottomrule
    \end{tabular*}
    \vspace{-10pt}
\end{table}

\noindent\textbf{Experimental setup.}
We further evaluate A-FAR on a physical active-camera manipulation platform to assess its performance under real-world visibility disruptions. As illustrated in Fig.~\ref{fig:real_robot_setup}(a), our platform consists of a 6-DoF CamBot observation arm equipped with an Intel RealSense D435i camera and a 6-DoF Piper manipulation arm. We consider three manipulation tasks, including Button Pressing, Pepper Picking, and Rack Cleaning. For each task, we collect 50 VR-teleoperated demonstrations under Stage Occlusion for policy training. We compare A-FAR against ACT, $\pi_{0.5}$, and ManiFlow under both Clean and Stage Occlusion conditions, with 10 evaluation trials per method, task, and condition.

\noindent\textbf{Quantitative results.}
As reported in Table~\ref{tab:real_world_results}, A-FAR achieves the highest SR on all three tasks under Stage Occlusion, reaching an average of 73.3\% compared with 43.3\% for the strongest baseline. A-FAR also leads under Clean with an average SR of 66.7\%. These results demonstrate that A-FAR maintains strong manipulation performance on physical hardware, particularly when task-relevant visual information is disrupted by external occlusions.

\noindent\textbf{Qualitative observations.}
As illustrated in Fig.~\ref{fig:real_robot_setup}(b), A-FAR actively adjusts the observation arm to restore task-relevant visibility after external occlusions, enabling subsequent manipulation to continue. This behavior demonstrates that a unified visuomotor policy can coordinate physical viewpoint recovery with downstream manipulation without requiring a separate perception or viewpoint planning module.

\section{Conclusion}
\label{sec:conclusion}

We study physical active vision as closed-loop recovery from external visibility disruptions during manipulation. We introduce BAVO-Bench, which systematically evaluates active-view recovery under controlled visibility conditions, together with CVG to isolate visibility gains attributable to camera motion. We further propose A-FAR, which combines robot-centric 3D observations, spatial context modeling, and future 3D motion supervision for joint viewpoint and manipulation control. Experiments in simulation and the real world demonstrate improved manipulation performance and active-view recovery, while ablation and probing analyses support the benefits of viewpoint-stable and future-aware geometric representations. Our findings further reveal a visibility-execution gap, underscoring the importance of translating recovered visual information into effective downstream manipulation control.

\section*{LLM Usage Statement}

We used OpenAI ChatGPT and Codex to assist with author-led research and manuscript preparation. ChatGPT supported literature discovery and summarization, facilitated discussions of experimental results, and assisted with manuscript drafting and refinement. Codex was used for code development, debugging, and implementation refinement. These tools served as research assistants rather than autonomous decision-makers, with research directions, methodological decisions, and scientific conclusions determined by the authors.

All AI-assisted materials were critically reviewed and revised by the authors before inclusion. Literature references and factual claims were checked against original sources, experimental results and tabulated values were verified against underlying experimental records, and manuscript text was reviewed for accuracy, consistency, and fidelity to the intended scientific claims. Unsupported or inaccurate AI-generated suggestions were corrected or excluded. AI outputs were not treated as independent scientific evidence. The authors retained full responsibility for research design, implementation, data collection, evaluation, and scientific interpretation, as well as the accuracy, originality, and integrity of the final manuscript and reported results.

{\small
    \bibliographystyle{ieeenat_fullname}
    \bibliography{references}
}

\clearpage
\maketitlesupplementary

This supplementary material provides additional benchmark specifications, implementation details, extended experimental analyses, real-world platform documentation, and a discussion of limitations.

\begin{itemize}
    \item \textbf{Appendix~\ref{app:benchmark_details}} details the BAVO-Bench task and milestone definitions, occlusion protocols, evaluation metric implementations, and representative rollout visualizations.
    \item \textbf{Appendix~\ref{app:method_details}} describes the robot-centric 3D preprocessing pipeline, OpenD4RT future motion supervision, A-FAR network architecture, training hyperparameters, and baseline configurations.
    \item \textbf{Appendix~\ref{app:additional_results}} presents complete per-task simulation results and details for the future visibility probing analysis.
    \item \textbf{Appendix~\ref{app:real_world}} documents the physical active-vision hardware platform, calibration and depth processing, real-world evaluation protocols, and rollout visualizations.
    \item \textbf{Appendix~\ref{app:limitations}} discusses current limitations and directions for future work.
\end{itemize}


\appendix
\section{BAVO-Bench Details}
\label{app:benchmark_details}

\subsection{Environment and Task Definitions}

BAVO-Bench builds on the AV-ALOHA simulator at 25\,Hz, with evaluation episodes capped at 500 control steps and terminating early upon success.
Under joint-target control, both the state and action spaces are 21-dimensional across the two manipulation arms with grippers and the 7-DoF active observation arm, while visual input comes solely from the single active RGB-D camera.
Task-relevant object poses are randomized at reset according to task-specific workspace layouts.
The benchmark contains five multi-stage tasks---\textit{Cube Handoff}, \textit{Block Stacking}, \textit{Drawer Stowing}, \textit{Slide Retrieval}, and \textit{Dustpan Sweeping}---each decomposed into five ordered milestones $M_1$ through $M_5$, with $M_5$ defining task success.
Milestones are evaluated from privileged simulator states and contact predicates for benchmarking and occlusion scheduling, and are never exposed to the policy.
Table~\ref{tab:task_milestones} summarizes the task objectives and milestone progression.

\subsection{Occlusion Protocol}

Both occlusion conditions introduce external visibility disruptions while leaving the underlying manipulation task unchanged.
Once placed, the occluder remains fixed in the world frame until the next intervention, so recovering visibility requires moving the active observation camera.

\noindent\textbf{Stage Occlusion.}
Each task defines three stage-dependent occlusion targets, starting with an initial occlusion at episode reset and updating at two predefined milestone transitions, as summarized in Table~\ref{tab:stage_occlusion}.
The occlusion target is determined directly by the current task stage rather than policy behavior.

\noindent\textbf{Random-time Occlusion.}
Random-time Occlusion uses the same stage-dependent target mapping but decouples subsequent interventions from milestone transitions.
Beyond the initial occlusion at reset, one trigger step is sampled per episode from each of six control-step windows,
\begin{equation*}
\begin{gathered}
\left[60,130\right],\ \left[130,200\right],\ \left[200,270\right],\\
\left[270,340\right],\ \left[340,410\right],\ \left[410,480\right],
\end{gathered}
\end{equation*}
with consecutive triggers separated by more than 60 steps.
When triggered, the target corresponds to the task stage reached at that step, allowing disruptions to interrupt ongoing manipulation.

\noindent\textbf{Occluder placement.}
Both settings use a camera-facing thin box placed along the line of sight between the active camera and the current target.
A candidate placement is accepted if it leaves at most two visible target pixels in the segmentation render while maintaining at least $1.5\,\text{cm}$ geometric clearance from the robot arms in MuJoCo.

\begin{table*}[t]
\caption{\textbf{Task objectives and milestone progression in BAVO-Bench.}
Each task contains five milestones, with $M_5$ defining task success.}
\label{tab:task_milestones}
\vspace{-10pt}
\centering
\small
\setlength{\tabcolsep}{4pt}
\renewcommand{\arraystretch}{1.08}
\begin{tabular}{
@{}
>{\raggedright\arraybackslash}p{0.15\linewidth}
>{\raggedright\arraybackslash}p{0.23\linewidth}
>{\raggedright\arraybackslash}p{0.58\linewidth}
@{}
}
\toprule
\textbf{Task} & \textbf{Objective} & \textbf{Ordered milestones} \\
\midrule

\textbf{Cube Handoff} &
Hand off a cube from the right arm to the left arm and place it on the target plate. &
$M_1$: the right gripper contacts the cube;
$M_2$: the right arm lifts the cube;
$M_3$: both grippers hold the cube during handoff;
$M_4$: the left gripper takes over and the right gripper releases;
$M_5$: the left arm places the cube on the target plate. \\
\addlinespace[4pt]

\textbf{Block Stacking} &
Move the base block to the central workspace and stack the top block on it. &
$M_1$: the right gripper contacts the base block;
$M_2$: the base block is placed in the stacking region and released;
$M_3$: the left gripper contacts the top block;
$M_4$: the top block is lifted into the pre-stacking region;
$M_5$: a stable stack is formed with both grippers released. \\
\addlinespace[4pt]

\textbf{Drawer Stowing} &
Open the drawer, stow an object inside, and close the drawer. &
$M_1$: the right gripper contacts the drawer handle;
$M_2$: the drawer is opened;
$M_3$: the left gripper contacts the object;
$M_4$: the object is placed and released inside the drawer;
$M_5$: the drawer is closed with the object remaining inside. \\
\addlinespace[4pt]

\textbf{Slide Retrieval} &
Slide the handled flat tool to expose its handle, then retrieve it with the left arm. &
$M_1$: the right gripper contacts the blade;
$M_2$: the handle is exposed by sliding;
$M_3$: the right gripper releases the tool;
$M_4$: the left gripper contacts the exposed handle;
$M_5$: the left arm lifts the tool from the support surface. \\
\addlinespace[4pt]

\textbf{Dustpan Sweeping} &
Position the dustpan and sweep the target object into it. &
$M_1$: the left gripper contacts the dustpan handle;
$M_2$: the dustpan is positioned in a valid sweeping configuration and released;
$M_3$: the right gripper contacts the broom;
$M_4$: the broom contacts the target object during sweeping;
$M_5$: the target object reaches the dustpan entrance. \\

\bottomrule
\end{tabular}
\end{table*}

\begin{table}[t]
\caption{\textbf{Stage Occlusion schedule.}
An initial occlusion is introduced at reset, followed by two stage-aligned target updates.}
\label{tab:stage_occlusion}
\vspace{-10pt}
\centering
\footnotesize
\setlength{\tabcolsep}{4pt}
\renewcommand{\arraystretch}{1.15}
\resizebox{\linewidth}{!}{%
\begin{tabular}{@{}llcl@{}}
\toprule
\textbf{Task} & \textbf{Initial target} & \textbf{Transitions} & \textbf{Subsequent targets} \\
\midrule
Cube Handoff & cube & $M_2, M_4$ & cube $\rightarrow$ plate \\
Block Stacking & base block & $M_2, M_4$ & top block $\rightarrow$ base block \\
Drawer Stowing & drawer front & $M_2, M_3$ & object $\rightarrow$ drawer front \\
Slide Retrieval & support platform & $M_1, M_2$ & blade $\rightarrow$ handle \\
Dustpan Sweeping & dustpan & $M_2, M_3$ & broom $\rightarrow$ target object \\
\bottomrule
\end{tabular}%
}
\vspace{-10pt}
\end{table}

\subsection{Metric Implementation Details}

\noindent\textbf{Task progress.}
Milestone progress for TPS is recorded monotonically, so reaching a milestone fixes the minimum progress for the episode regardless of subsequent state changes.
As defined in Sec.~\ref{sec:benchmark_metrics}, reaching $M_5$ defines task success.

\noindent\textbf{CVG evaluation.}
CVG is evaluated under Random-time Occlusion over both the initial occlusion at reset and all subsequent triggered events.
In Eq.~\ref{eq:event_cvg}, we use a recovery horizon of $H=60$ control steps (2.4\,s at 25\,Hz), so recovery windows from consecutive events do not overlap.
Visible target pixel counts $N^{\mathrm{occ}}_e$ and $N^{\mathrm{noOcc}}_e$ are obtained from the segmentation render; when $N^{\mathrm{noOcc}}_e=0$, normalized visibility is set to zero.
Events truncated by episode termination before the full window completes are excluded rather than padded.
For episodes with multiple retained events, we average event-level CVG within each episode before averaging across episodes.
Counterfactual views are rendered from the same simulator state by temporarily modifying only the evaluation camera pose and benchmark occluder, without affecting policy execution.

\subsection{Benchmark Rollout Visualizations}
\label{app:benchmark_visualizations}

\begin{figure}[t!]
    \centering
    \includegraphics[
        width=\linewidth,
        height=0.82\textheight,
        keepaspectratio
    ]{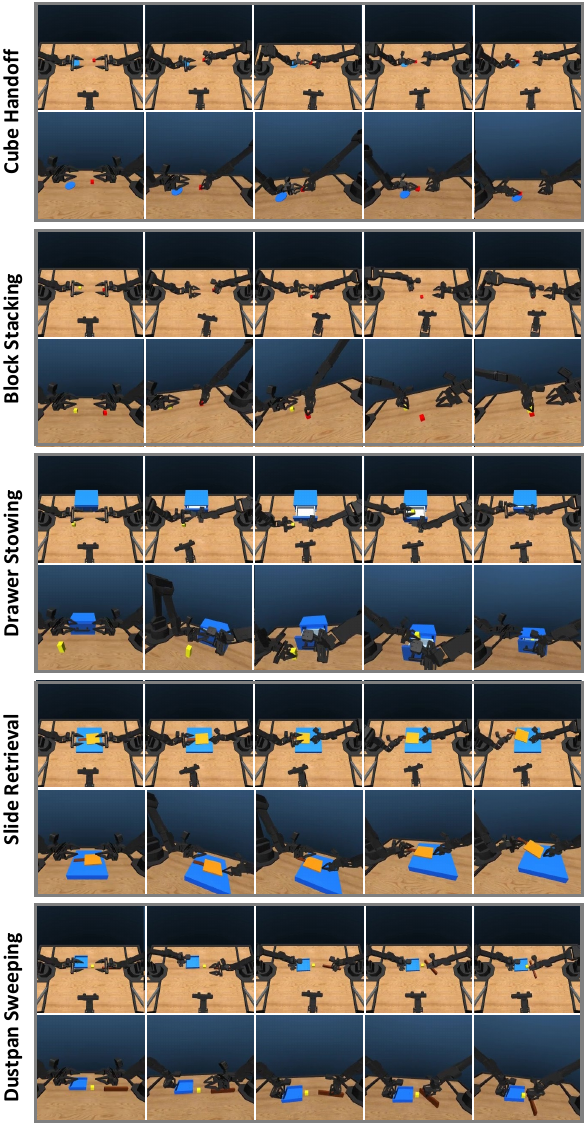}
    \caption{\textbf{Rollout visualizations under Clean.}
    Representative sequences for the five benchmark tasks without external
    visibility intervention.
    For each task, the top row shows the third-person scene view and the
    bottom row shows the active-camera view.}
    \label{fig:appendix_clean_rollouts}
\end{figure}

\begin{figure}[t!]
    \centering
    \includegraphics[
        width=\linewidth,
        height=0.82\textheight,
        keepaspectratio
    ]{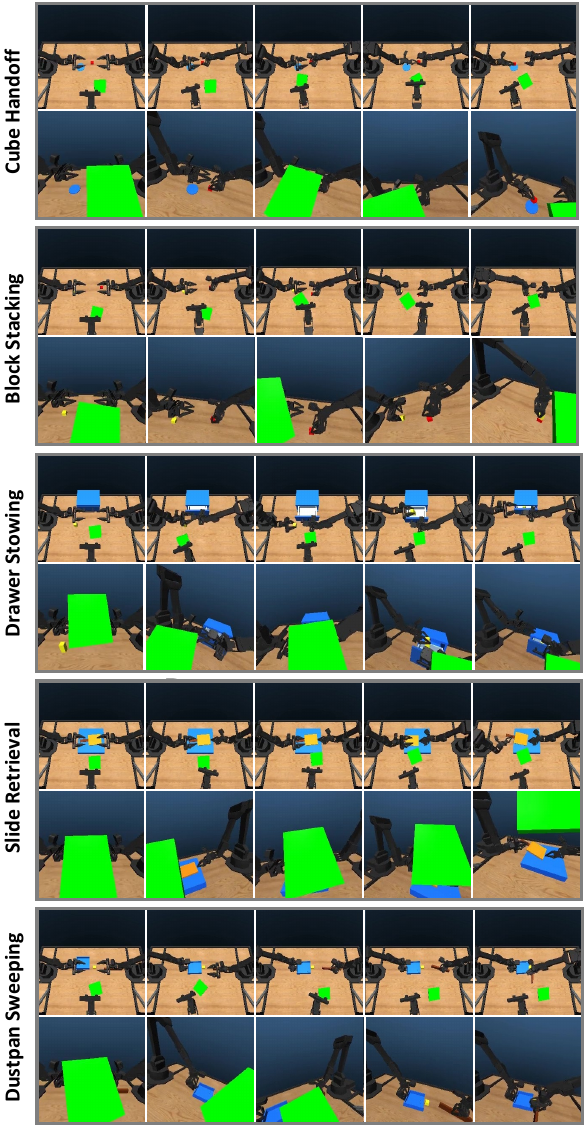}
    \caption{\textbf{Rollout visualizations under Stage Occlusion.}
    Representative sequences for the five benchmark tasks with stage-aligned
    occlusion events.
    For each task, the top row shows the third-person scene view and the
    bottom row shows the active-camera view.}
    \label{fig:appendix_stage_rollouts}
\end{figure}

\begin{figure}[t!]
    \centering
    \includegraphics[
        width=\linewidth,
        height=0.82\textheight,
        keepaspectratio
    ]{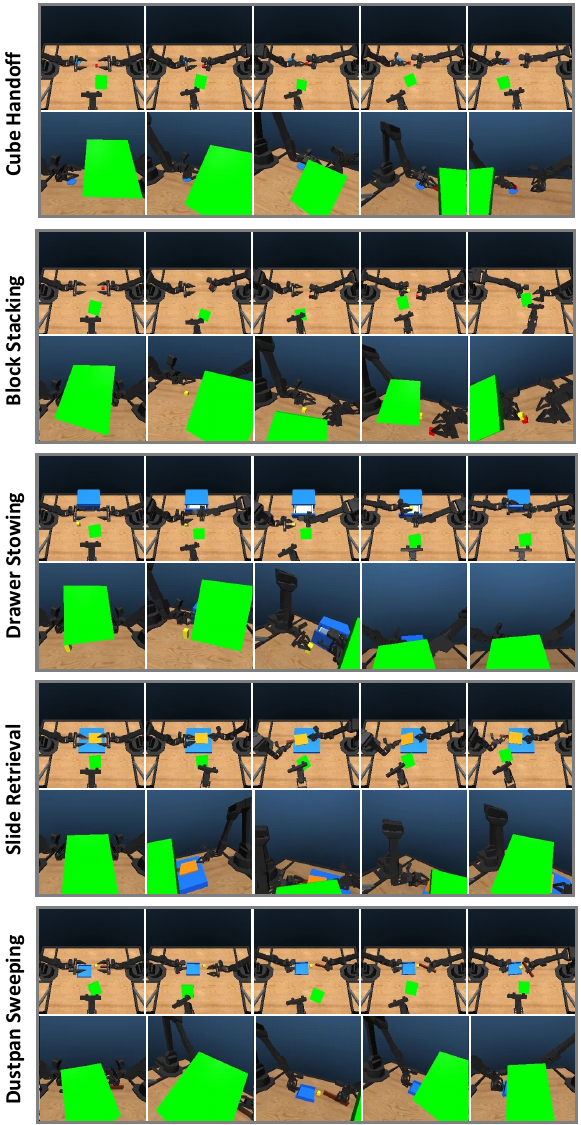}
    \caption{\textbf{Rollout visualizations under Random-time Occlusion.}
    Representative sequences for the five benchmark tasks with temporally
    unexpected occlusion events.
    For each task, the top row shows the third-person scene view and the
    bottom row shows the active-camera view.}
    \label{fig:appendix_random_rollouts}
\end{figure}

Figs.~\ref{fig:appendix_clean_rollouts}, \ref{fig:appendix_stage_rollouts}, and \ref{fig:appendix_random_rollouts} show representative rollouts across all five tasks under Clean, Stage Occlusion, and Random-time Occlusion, ordered from top to bottom as \textit{Cube Handoff}, \textit{Block Stacking}, \textit{Drawer Stowing}, \textit{Slide Retrieval}, and \textit{Dustpan Sweeping}.
For each task, the top row shows the third-person view and the bottom row shows the active-camera view from left to right.

\section{A-FAR and Implementation Details}
\label{app:method_details}

\subsection{3D Preprocessing and Motion Supervision}

\noindent\textbf{Point-cloud preprocessing.}
RGB-D observations are back-projected into 3D and transformed into the shared robot-base frame.
Points are cropped to the workspace bounding box and downsampled via farthest point sampling (FPS) to $N=256$ colored XYZRGB points without further subsampling inside the policy.

\noindent\textbf{OpenD4RT motion targets.}
We extract pointwise 3D motion offline using a frozen OpenD4RT teacher~\citep{opend4rt} on 32-frame RGB clips resized to $256\times256$.
Using zero-based indexing, frame 16 serves as the current anchor and frame 31 as the future target ($\Delta t=15$).
For each of the 256 sampled points, OpenD4RT predicts its 3D displacement, which is rotated into the robot-base frame.
A motion target is marked valid ($m_i=1$) when the source point is valid, the predicted displacement is finite with norm below $0.5\,\text{m}$, and the 15-step window lies within the episode without crossing a task stage transition.
Before computing the Smooth L1 loss, displacements are scaled by $0.1\,\text{m}$ and softly weighted by $0.3 + 0.7\,p_{\mathrm{vis},i}$ using the teacher visibility confidence so that temporarily occluded points retain supervision.

\subsection{Architecture and Training Configuration}

The ManiFlow PointNet encoder extracts 128-dimensional features per point.
The Spatial Context Module adds 3D positional embeddings generated from point coordinates and applies two self-attention blocks (4 heads, hidden dimension 128, FFN dimension 512).
During training, the auxiliary motion head $g_\psi$ encodes the 15-step demonstrated future action sequence into a 128-dimensional embedding, adds it to each contextualized point token, and predicts per-point 3D displacements via a shared MLP.
The 21-dimensional robot state is projected to a 64-dimensional embedding and randomly masked with probability $p=0.15$ once per training window, shared across the two observation frames.

A-FAR is trained for 100K steps with batch size 64, AdamW at learning rate $1\times10^{-4}$, and motion loss weight $\lambda_{\mathrm{motion}}=0.1$.
The policy takes an observation horizon of 2 frames, predicts a 16-step action horizon, and executes 15 open-loop steps during evaluation.
Table~\ref{tab:afar_training_config} summarizes the main architecture and training settings.

\begin{table}[t]
\caption{\textbf{Architecture and training settings for A-FAR.}}
\label{tab:afar_training_config}
\vspace{-10pt}
\centering
\small
\setlength{\tabcolsep}{6pt}
\begin{tabular}{@{}lc@{}}
\toprule
\textbf{Configuration} & \textbf{Value} \\
\midrule
Point-cloud size $N$ & 256 \\
Observation horizon & 2 \\
Action horizon / execution steps & 16 / 15 \\
Motion supervision horizon $\Delta t$ & 15 \\
Point / state feature dimension & 128 / 64 \\
Spatial Context layers / heads & 2 / 4 \\
Proprioceptive masking prob.\ $p$ & 0.15 \\
Motion loss weight $\lambda_{\mathrm{motion}}$ & 0.1 \\
Batch size & 64 \\
Optimizer / learning rate & AdamW / $1\times10^{-4}$ \\
Training steps & 100K \\
\bottomrule
\end{tabular}
\vspace{-10pt}
\end{table}

\subsection{Baseline Configurations}

\noindent\textbf{ACT.}
Trained per task from monocular active-camera RGB and the 21-dimensional robot state for 30K steps with batch size 64.
Following its standard protocol, ACT replans at every control step with temporal ensembling (weight 0.01).

\noindent\textbf{$\bm{\pi_{0.5}}$.}
Fine-tuned once jointly across all five tasks using monocular active-camera RGB resized to $224\times224$ for 50K steps with batch size 32, and evaluated with 15-step open-loop action execution.

\noindent\textbf{ManiFlow.}
Trained per task with the same robot-base 256-point observations, 21-dimensional robot state, batch size 64, 100K steps, and 15-step open-loop execution as A-FAR.
\section{Additional Experimental Results}
\label{app:additional_results}

We provide the complete per-task breakdown corresponding to the macro-averaged simulation results in the main paper, along with implementation details for the future visibility probing analysis.
Task-level CVG and Joint scores under Random-time Occlusion are already reported in Table~\ref{tab:active_view}.

\subsection{Per-Task Simulation Results}
\label{app:task_results}

Table~\ref{tab:appendix_per_task} reports per-task Success Rate (SR) and Task Progress Score (TPS) across Clean, Stage Occlusion, and Random-time Occlusion.
Under Stage Occlusion, A-FAR achieves the highest TPS on all five tasks and the highest SR on four of five tasks, with clear gains over ManiFlow on \textit{Block Stacking} (30.0\% vs.\ 13.3\%) and \textit{Drawer Stowing} (95.0\% vs.\ 83.3\%).
Under Clean observations, A-FAR improves both SR and TPS over ManiFlow on \textit{Cube Handoff}, \textit{Block Stacking}, and \textit{Slide Retrieval}, while ManiFlow remains stronger on \textit{Drawer Stowing} and \textit{Dustpan Sweeping}.
Under Random-time Occlusion, A-FAR improves SR over ManiFlow on all four tasks with non-zero success, whereas all methods remain at 0.0\% SR on \textit{Block Stacking} despite partial progress (32.0\% TPS for A-FAR).
ACT achieves higher Random-time SR on \textit{Drawer Stowing} and \textit{Slide Retrieval} under per-step replanning with temporal ensembling, but fails on the tighter bimanual coordination required in \textit{Cube Handoff} and \textit{Block Stacking}.

\begin{table*}[t]
\caption{\textbf{Per-task Success Rate (SR, \%) and Task Progress Score (TPS, \%) on BAVO-Bench.}
Results are reported under Clean, Stage Occlusion, and Random-time Occlusion.
ACT, ManiFlow, and A-FAR report mean$\pm$std over three training seeds; $\pi_{0.5}$ is evaluated from a single multi-task fine-tuning run.
Best performance for each task and metric is shown in bold (none bolded when all methods tie at $0.0$).}
\label{tab:appendix_per_task}
\vspace{-10pt}
\centering
\footnotesize
\setlength{\tabcolsep}{5pt}
\renewcommand{\arraystretch}{1.08}
\begin{tabular}{@{}llcccccc@{}}
\toprule
& & \multicolumn{2}{c}{\textbf{Clean}} & \multicolumn{2}{c}{\textbf{Stage Occlusion}} & \multicolumn{2}{c}{\textbf{Random-time Occlusion}} \\
\cmidrule(lr){3-4} \cmidrule(lr){5-6} \cmidrule(lr){7-8}
\textbf{Task} & \textbf{Method} & \textbf{SR} & \textbf{TPS} & \textbf{SR} & \textbf{TPS} & \textbf{SR} & \textbf{TPS} \\
\midrule

Cube Handoff
& ACT & $0.0{\pm}0.0$ & $33.0{\pm}2.6$ & $6.7{\pm}7.6$ & $44.0{\pm}3.6$ & $0.0{\pm}0.0$ & $24.0{\pm}6.2$ \\
& $\pi_{0.5}$ & $5.0$ & $45.0$ & $30.0$ & $68.0$ & $0.0$ & $31.0$ \\
& ManiFlow & $41.7{\pm}2.9$ & $70.3{\pm}5.0$ & $\mathbf{68.3{\pm}5.8}$ & $87.7{\pm}2.3$ & $8.3{\pm}7.6$ & $\mathbf{45.3{\pm}5.7}$ \\
& \textbf{A-FAR} & $\mathbf{50.0{\pm}13.2}$ & $\mathbf{84.0{\pm}4.6}$ & $63.3{\pm}2.9$ & $\mathbf{89.7{\pm}0.6}$ & $\mathbf{10.0{\pm}5.0}$ & $44.7{\pm}2.1$ \\
\midrule

Block Stacking
& ACT & $0.0{\pm}0.0$ & $24.3{\pm}3.5$ & $0.0{\pm}0.0$ & $28.7{\pm}2.9$ & $0.0{\pm}0.0$ & $22.3{\pm}2.1$ \\
& $\pi_{0.5}$ & $10.0$ & $69.0$ & $15.0$ & $69.0$ & $0.0$ & $22.0$ \\
& ManiFlow & $8.3{\pm}7.6$ & $44.7{\pm}20.3$ & $13.3{\pm}2.9$ & $62.7{\pm}6.7$ & $0.0{\pm}0.0$ & $30.3{\pm}3.2$ \\
& \textbf{A-FAR} & $\mathbf{40.0{\pm}13.2}$ & $\mathbf{81.7{\pm}5.5}$ & $\mathbf{30.0{\pm}8.7}$ & $\mathbf{79.7{\pm}3.2}$ & $0.0{\pm}0.0$ & $\mathbf{32.0{\pm}5.3}$ \\
\midrule

Drawer Stowing
& ACT & $50.0{\pm}5.0$ & $78.7{\pm}3.1$ & $56.7{\pm}2.9$ & $80.7{\pm}2.1$ & $\mathbf{45.0{\pm}8.7}$ & $\mathbf{77.3{\pm}4.7}$ \\
& $\pi_{0.5}$ & $25.0$ & $70.0$ & $75.0$ & $90.0$ & $0.0$ & $51.0$ \\
& ManiFlow & $\mathbf{90.0{\pm}13.2}$ & $\mathbf{96.0{\pm}5.3}$ & $83.3{\pm}5.8$ & $93.3{\pm}2.3$ & $38.3{\pm}14.4$ & $65.3{\pm}5.5$ \\
& \textbf{A-FAR} & $81.7{\pm}7.6$ & $92.0{\pm}3.5$ & $\mathbf{95.0{\pm}5.0}$ & $\mathbf{97.3{\pm}2.3}$ & $41.7{\pm}5.8$ & $69.0{\pm}0.0$ \\
\midrule

Slide Retrieval
& ACT & $0.0{\pm}0.0$ & $48.0{\pm}19.0$ & $31.7{\pm}15.3$ & $65.7{\pm}9.5$ & $\mathbf{30.0{\pm}8.7}$ & $62.7{\pm}3.8$ \\
& $\pi_{0.5}$ & $0.0$ & $40.0$ & $55.0$ & $74.0$ & $10.0$ & $53.0$ \\
& ManiFlow & $33.3{\pm}12.6$ & $63.7{\pm}7.0$ & $58.3{\pm}2.9$ & $74.3{\pm}2.5$ & $13.3{\pm}2.9$ & $58.0{\pm}3.5$ \\
& \textbf{A-FAR} & $\mathbf{51.7{\pm}11.5}$ & $\mathbf{73.3{\pm}3.8}$ & $\mathbf{68.3{\pm}2.9}$ & $\mathbf{79.7{\pm}1.5}$ & $23.3{\pm}7.6$ & $\mathbf{63.3{\pm}3.5}$ \\
\midrule

Dustpan Sweeping
& ACT & $1.7{\pm}2.9$ & $47.0{\pm}17.3$ & $28.3{\pm}16.1$ & $71.3{\pm}8.6$ & $3.3{\pm}5.8$ & $46.7{\pm}2.5$ \\
& $\pi_{0.5}$ & $25.0$ & $52.0$ & $35.0$ & $66.0$ & $10.0$ & $47.0$ \\
& ManiFlow & $\mathbf{55.0{\pm}15.0}$ & $\mathbf{77.7{\pm}3.1}$ & $80.0{\pm}8.7$ & $89.7{\pm}5.0$ & $15.0{\pm}5.0$ & $\mathbf{53.7{\pm}6.5}$ \\
& \textbf{A-FAR} & $26.7{\pm}2.9$ & $46.7{\pm}4.0$ & $\mathbf{91.7{\pm}10.4}$ & $\mathbf{94.7{\pm}7.6}$ & $\mathbf{26.7{\pm}5.8}$ & $51.7{\pm}4.0$ \\

\bottomrule
\end{tabular}
\end{table*}

\subsection{Future Visibility Probing Details}
\label{app:probing_details}

For the representation probing experiment in Sec.~\ref{sec:representation_analysis} of the main paper, we freeze each policy's visual encoder and extract 128-dimensional pointwise features.
Following the 80/20 episode split described in the main paper, each probe is trained for 2,000 steps with batch size 16 using positive-weighted binary cross-entropy loss and AdamW at learning rate $1\times10^{-3}$.

Ground-truth visibility labels at $t+15$ are generated from simulator geometry by tracking each observed 3D surface point to frame $t+15$.
A point is labeled visible if its projection lies within the image bounds and a rendered pixel in its $3\times3$ neighborhood shares the same geometry ID with depth error at most $5\,\text{mm}$; points outside the view or occluded are labeled invisible, and unmatched points are excluded.
\section{Real-World System and Experiments}
\label{app:real_world}

\subsection{Hardware Platform and 3D Preprocessing}

\noindent\textbf{Platform and control.}
The physical platform pairs a 6-DoF Piper manipulation arm (with a parallel gripper) and a 6-DoF CamBot observation arm carrying a RealSense D435i RGB-D camera.
Under joint-target control, both the proprioceptive state and action spaces are 13-dimensional, combining 6 joint positions and 1 gripper state for the manipulation arm with 6 joint positions for the observation arm.

\noindent\textbf{Calibration and point-cloud pipeline.}
We perform checkerboard hand-eye calibration to estimate the fixed camera-to-end-effector transform $T_{E\leftarrow C}$, and compute the time-varying end-effector-to-base transform $T_{B\leftarrow E_t}$ via CamBot forward kinematics.
Their composition $T_{B\leftarrow C_t} = T_{B\leftarrow E_t}T_{E\leftarrow C}$ maps active-camera observations into the stationary observation-arm base frame.
To handle real-world depth noise and background clutter, we first mask out green tabletop pixels and invalid depths in image space before back-projection.
The remaining pixels are lifted to 3D, transformed into the base frame, cropped to the workspace, denoised with DBSCAN to remove depth outliers, and downsampled via FPS to $N=256$ colored points.

\subsection{Tasks, Occlusion Protocol, and Training}

\noindent\textbf{Tasks and Stage Occlusion.}
We evaluate three single-arm active-vision tasks with randomized initial object placements across the workspace.
During Stage Occlusion demonstrations and evaluation trials, a human operator places a handheld occluder along the camera's line of sight at predefined task phases consistently across all methods:
\begin{itemize}
    \item \textit{Pepper Picking}: grasp a pepper and place it into a bowl; two occlusions are introduced, first prior to grasping the pepper and again prior to placing it into the bowl.
    \item \textit{Button Pressing}: press the target button on the workspace; one occlusion is introduced before the arm approaches the button.
    \item \textit{Rack Cleaning}: grasp a cloth and wipe across a rack from right to left and back; two occlusions are introduced, first prior to grasping the cloth and again when the cloth reaches the left side of the rack.
\end{itemize}

\noindent\textbf{Training and evaluation.}
Policies are trained on 50 VR-teleoperated Stage Occlusion demonstrations per task with the same training steps and action execution horizons as in simulation (Appendix~\ref{app:method_details}).
Consistent with our simulation protocol, ACT, ManiFlow, and A-FAR are trained per task, whereas $\pi_{0.5}$ is jointly fine-tuned once across all three real-world tasks.

\subsection{Real-World Rollout Visualizations}
\label{app:real_world_visualizations}

\begin{figure*}[p]
    \centering
    \includegraphics[
        width=\textwidth,
        height=0.40\textheight,
        keepaspectratio
    ]{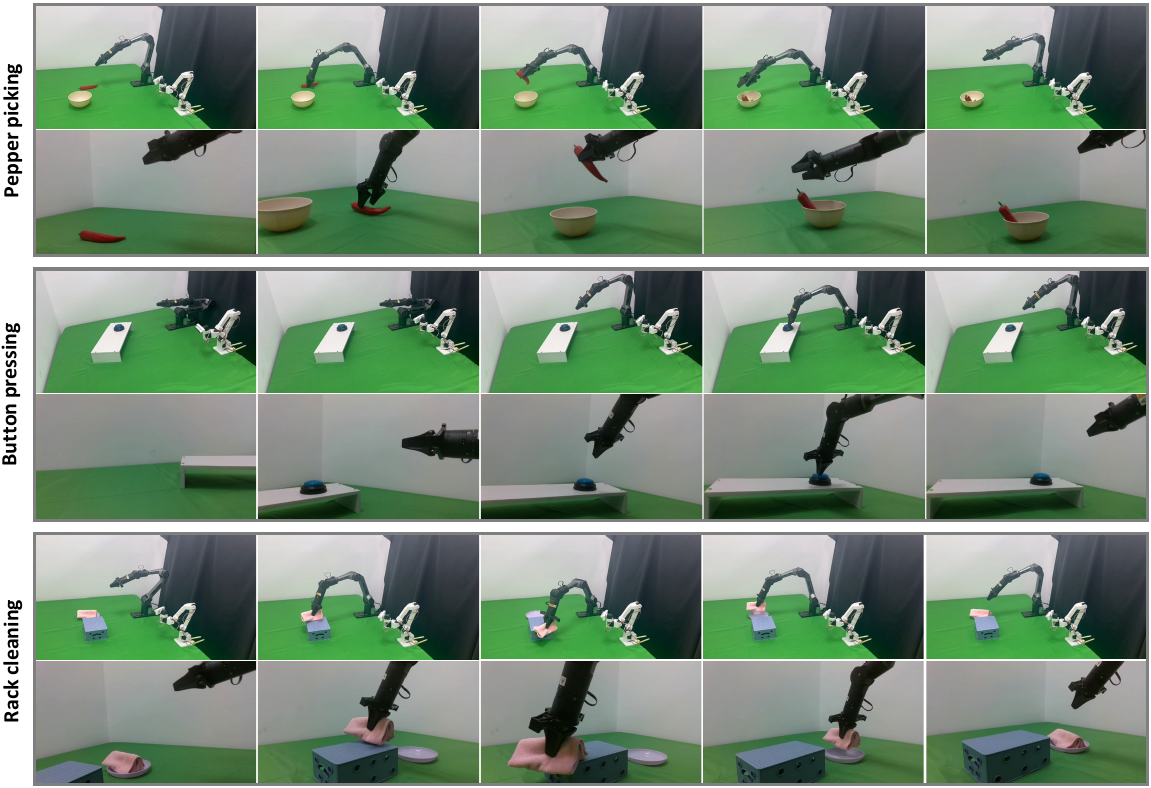}
    \caption{\textbf{Real-world rollout visualizations under Clean.}
    Representative sequences for the three physical tasks without external occlusion.
    For each task, the top row shows the third-person scene view and the bottom row shows the active-camera view from left to right.}
    \label{fig:appendix_real_clean}

    \vspace{12pt}

    \includegraphics[
        width=\textwidth,
        height=0.40\textheight,
        keepaspectratio
    ]{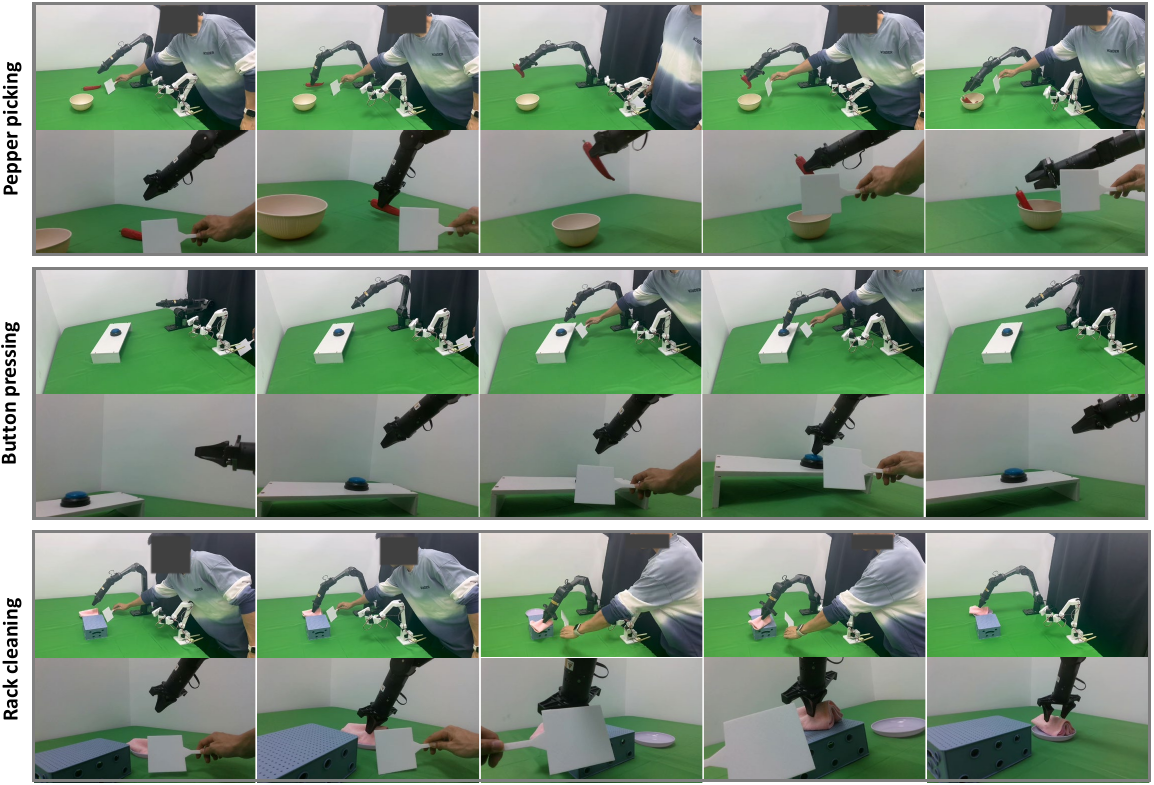}
    \caption{\textbf{Real-world rollout visualizations under Stage Occlusion.}
    Representative sequences for the three physical tasks under human-introduced occlusions, illustrating active viewpoint adjustment and continued manipulation.
    For each task, the top row shows the third-person scene view and the bottom row shows the active-camera view from left to right.}
    \label{fig:appendix_real_stage}
\end{figure*}

Figs.~\ref{fig:appendix_real_clean} and~\ref{fig:appendix_real_stage} show representative rollouts on our physical platform under Clean and Stage Occlusion, ordered from top to bottom as \textit{Pepper Picking}, \textit{Button Pressing}, and \textit{Rack Cleaning}.
For each task, the top row shows the third-person view and the bottom row shows the active-camera view from left to right.

\clearpage

\section{Limitations and Future Work}
\label{app:limitations}

\noindent\textbf{Rapid recovery under precision constraints.}
Under frequent, unscheduled disruptions in Random-time Occlusion, policies must restore visibility rapidly without destabilizing ongoing manipulation.
Precision-critical tasks such as \textit{Cube Handoff} and \textit{Block Stacking} remain particularly challenging, where open-loop action chunking introduces reaction latency and mid-trajectory interruptions still break fine alignment despite substantial visibility recovery.
Future work can explore higher-frequency closed-loop replanning, asynchronous or decoupled viewpoint-manipulation control, and short-term 3D scene memory to preserve manipulation precision during transient occlusions.

\noindent\textbf{Learning paradigms beyond in-distribution imitation.}
Training solely via offline imitation from limited Stage Occlusion demonstrations is most effective for occlusion patterns close to the training distribution, and generalizing to arbitrary timing, viewpoints, and occluder geometries remains difficult.
Achieving broadly robust active-view recovery likely requires new learning paradigms beyond narrow behavior cloning.
Two promising directions are scaling diverse active-vision data within generalist Vision-Language-Action pretraining so reactive recovery emerges across tasks, and using large-scale reinforcement learning in simulation over diverse occluder shapes, motions, and disruption patterns followed by sim-to-real transfer.

\noindent\textbf{Benchmark scope and occlusion diversity.}
BAVO-Bench currently focuses on five tabletop bimanual manipulation tasks with a single quasi-static rigid occluder per intervention.
Expanding the benchmark to a broader set of manipulation tasks, longer-horizon workflows, and richer visibility disruptions such as continuously moving obstacles or multi-object occlusions would provide a more comprehensive testbed for physical active vision in unstructured environments.

\end{document}